\documentclass[10pt,twocolumn,letterpaper]{article}

\usepackage[pagenumbers]{cvpr} 
\usepackage{caption}
\usepackage{subcaption}
\usepackage{graphicx}
\usepackage{amsmath}
\usepackage{amssymb}
\usepackage{booktabs}
\usepackage{enumitem}

\usepackage{pifont}

\usepackage[table]{xcolor}
\usepackage[pagebackref, breaklinks=true, colorlinks, citecolor=citecolor, linkcolor=linkcolor, bookmarks=false]{hyperref}
\definecolor{citecolor}{HTML}{0071BC}
\definecolor{linkcolor}{HTML}{ED1C24}

\usepackage{algorithm}
\usepackage{bm}
\usepackage{algorithmic}
\usepackage{listings}
\usepackage{makecell}  
\usepackage{diagbox} 
\usepackage{mathtools} 
\usepackage{graphicx} 
\usepackage{multirow}
\usepackage{cleveref}

\def\confName{CVPR}
\def\confYear{2023}

\begin{document}
\title{Modulating CNN Features with Pre-Trained ViT Representations for Open-Vocabulary Object Detection}

\author{Xiangyu Gao, \quad Yu Dai, \quad Benliu Qiu, \quad Lanxiao Wang,\quad Heqian Qiu, \quad Hongliang Li\\
University of Electronic Science and Technology of China \\
{\tt\small \{xygao, ydai, qbenliu\}@std.uestc.edu.cn \quad  \{lanxiaowang, hqqiu, hlli\}@uestc.edu.cn}
}
\maketitle

\begin{abstract}
	Owing to large-scale image-text contrastive training, pre-trained vision language model (VLM) like CLIP shows superior open-vocabulary recognition ability. Most existing open-vocabulary object detectors attempt to utilize the pre-trained VLMs to attain generalized representation. F-ViT uses the pre-trained visual encoder as the backbone network and freezes it during training. However, its frozen backbone doesn't benefit from the labeled data to strengthen the representation for detection. Therefore, we propose a novel two-branch backbone network, named as \textbf{V}iT-Feature-\textbf{M}odulated Multi-Scale \textbf{C}onvolutional Network (VMCNet), which consists of a trainable convolutional branch, a frozen pre-trained ViT branch and a VMC module. The trainable CNN branch could be optimized with labeled data while the frozen pre-trained ViT branch could keep the representation ability derived from large-scale pre-training. Then, the proposed VMC module could modulate the multi-scale CNN features with the representations from ViT branch. With this proposed mixed structure, the detector is more likely to discover objects of novel categories.
	Evaluated on two popular benchmarks, our method boosts the detection performance on novel category and outperforms state-of-the-art methods. On OV-COCO, the proposed method achieves 44.3 AP$_{50}^{\mathrm{novel}}$ with ViT-B/16 and 48.5 AP$_{50}^{\mathrm{novel}}$ with ViT-L/14. On OV-LVIS, VMCNet with ViT-B/16 and ViT-L/14 reaches 27.8 and 38.4 mAP$_{r}$.
\end{abstract}

\section{Introduction}
\label{sec:intro}

Open-vocabulary object detection~\cite{DBLP:conf/cvpr/ZareianRHC21} (OVOD) requires models to detect the novel targets beyond the training-category set via the language instruction. Compared to traditional closed-set detector, OVOD model improves the generalization ability from instance-level to category-level, which makes a further step in the real-world applications such as autonomous driving, intelligent androids, medical imaging analysis, \etc.

\begin{figure*}
	\centering
	\subfloat[F-ViT Design~\cite{DBLP:conf/iclr/WuZX0L0L24}]{\includegraphics[width=0.2\textwidth,height=0.25\linewidth]{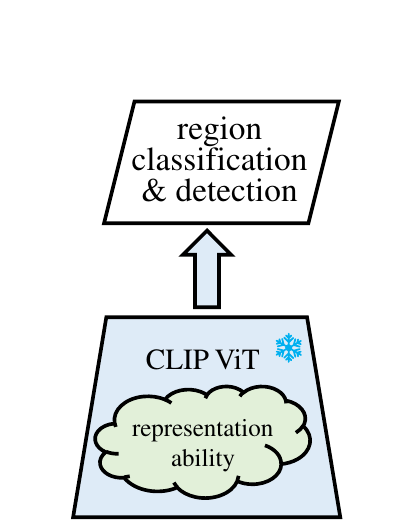}\label{fig:intro_a}}
	\hspace{0.035\textwidth}
	\subfloat[Parallel Design]{\includegraphics[width=0.28\textwidth,height=0.25\linewidth]{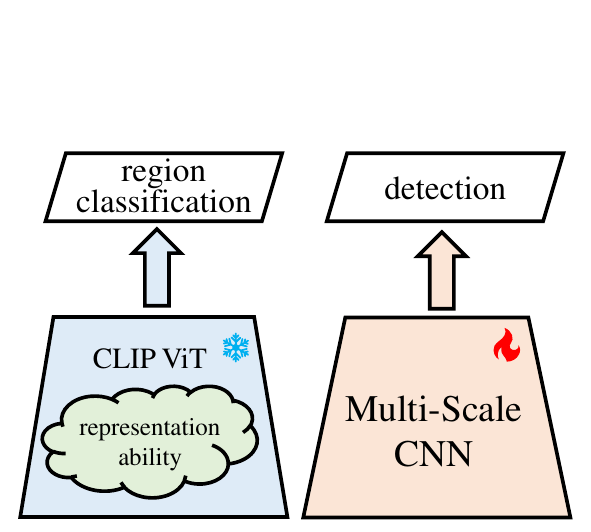}\label{fig:intro_b}}
	\hspace{0.035\textwidth}
	\subfloat[Ours]{\includegraphics[width=0.28\textwidth,height=0.25\linewidth]{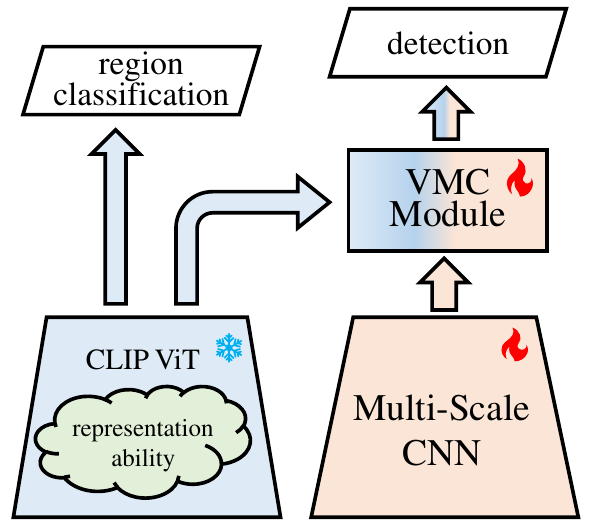}\label{fig:intro_c}}
	\caption{Different backbone paradigms for open-vocabulary object detection. Sign "snowflake" means this part is frozen during base training while "fire" represents trainable part. (a) F-ViT~\cite{DBLP:conf/iclr/WuZX0L0L24} uses a frozen CLIP ViT to extract image features, which does not employ the base training data to strengthen the representation for detection. (b) This paradigm uses an extra trainable convolutional neural network, which is optimized with base training data for detection. However, the representation ability of CLIP ViT is not exploited. (c) Our design applies two-branch architecture, the representations from frozen CLIP ViT are utilized to modulate the features from trainable CNN. Thus, the final representations for detection benefit from both the pre-trained model and base training data.}
	\label{fig:intro}
\end{figure*}

For most existing OVOD methods, pre-trained VLM is essential to implement open-vocabulary recognition. CLIP~\cite{DBLP:conf/icml/RadfordKHRGASAM21}, a typical VLM pre-trained on an enormous number of image-text pairs, is able to encode an image into the feature aligned with the text embedding space. To better satisfy the dense prediction task such as object detection, some works~\cite{DBLP:conf/eccv/ZhouLD22, DBLP:conf/cvpr/ZhongYZLCLZDYLG22, DBLP:conf/iclr/WuZX0L0L24, DBLP:journals/corr/abs-2404-08181} further improve CLIP, which transfer the visual encoder's representation ability from image-level to region-level. 
Based on the dense visual feature encoder, F-ViT~\cite{DBLP:conf/iclr/WuZX0L0L24} firstly builds an open-vocabulary object detector upon a frozen CLIP ViT backbone and achieves remarkable performance. 
Specifically, the ViT from CLIP is fine-tuned in a self-distillation manner at first. Then, F-ViT freezes this ViT during base training and leverages it as the backbone network to support detection.
Sketch of the feature flow in F-ViT is shown in~\cref{fig:intro_a}.

It is common knowledge that backbone network plays a key role in detection performance. In most detector designs, the backbone networks are optimized with the labeled bounding boxes during training, which ensures the detection performance. 
However, to save the generalization ability, the backbone network of F-ViT fails to exploit the knowledge of labeled data during base training. 
Although its self-distillation strategy strengthens the local representation of CLIP ViT which may implicitly promote detection, this strategy treats the vision transformer more like a region classifier than a backbone network. As the feature extractor, backbone network needs to suit the downstream modules in a detector via the bounding box training. Thus, merely using self-distillation may not be adequate for feature extraction of detection.

A usual scheme in object detection is to initialize the backbone with the weights from pre-training and fine-tune it with training data. However, directly fine-tuning the pre-trained model with limited data will harm its open-vocabulary recognition ability, which degrades the model into a closed-set detector. As reported in~\cite{DBLP:conf/iclr/WuZX0L0L24}, the performance on novel categories will degenerate when using trainable ViT backbone.
Another straightforward scheme is applying an extra trainable network to extract features parallelly for detection as shown in~\cref{fig:intro_b}.  Although the additional network exploits the training data, such design isolates the CLIP ViT, which does not leverage the representation ability of VLM for detection. 
Hence, it is nature to ask a question: could a backbone structure utilize the information from both the base training data and the pre-trained VLM simultaneously?

Inspired by ViT-CoMer~\cite{DBLP:conf/cvpr/XiaWLHS24}, we propose a two-branch backbone network, which could make use of the base training and pre-trained CLIP ViT.
The proposed backbone network, named as \textbf{V}iT-Feature-\textbf{M}odulated Multi-Scale \textbf{C}onvolutional Network (VMCNet), consists of a vision transformer branch, a multi-scale convolutional branch and a VMC module. \cref{fig:intro_c} shows the sketch of our design.
Firstly, to exploit the base training, we use a trainable convolutional neural network (CNN) as the convolutional branch. This CNN branch has a simple structure and is able to produce multi-scale features. Optimized with labeled bounding box, this branch could learn from base training data and provide multi-scale detection features. 
Besides, we need the representations from pre-trained VLM to strengthen the generalization ability of features. To reach this goal, we also freeze the CLIP ViT and use it as the second branch in our network. These ViT features are treated as intermediate products instead of the final output of backbone. 
Finally, given the features from two branches, the proposed VMC module modulates the multi-scale CNN features with the adaptive representations from ViT. By using the proposed structure, two branches are complementary to each other, and the knowledge from both the base training data and pre-trained ViT could be exploited to generate the detection features.
As a result, VMCNet improves the detection performance of OVOD detector on novel categories.

In summary, our main contributions are listed as follow:
\begin{itemize}
	\item We propose a novel backbone network for open-vocabulary object detector, named as \textbf{V}iT-Feature-\textbf{M}odulated Multi-Scale \textbf{C}onvolutional Network (VMCNet). 
	Via utilizing two-branch architecture, it leverages the strengths of different networks and exploits the information from both the base training data and the pre-trained VLM.
	
	\item We propose the VMC module to fuse the features from different branches. 
	By this module, multi-scale features from trainable convolutional network are modulated with the adapted features from frozen ViT. This feature fusion processing improves the quality of representation and boosts the detection performance on novel categories.
	
	\item We evaluate our proposed VMCNet on two popular benchmarks. The experimental results demonstrate that our method significantly improves the detection performance on novel categories. On OV-COCO, our method with ViT-B/16 achieves 44.3 AP$_{50}^{\mathrm{novel}}$. When utilizing ViT-L/14, VMCNet could achieve 48.5 AP$_{50}^{\mathrm{novel}}$. Moreover, VMCNet with ViT-B/16 and ViT-L/14 attains 27.8 and 38.4 mAP$_{r}$ on OV-LVIS, which surpasses state-of-the-art methods.
\end{itemize}

\section{Related Work}

\textbf{Backbone Structure for Object Detection.} In deep vision tasks, deep neural network was firstly applied in image classification. Typical CNNs such as AlexNet~\cite{DBLP:conf/nips/KrizhevskySH12}, VGGNet~\cite{DBLP:journals/corr/SimonyanZ14a} and ResNet~\cite{DBLP:conf/cvpr/HeZRS16} strongly promoted the development of vision tasks.
R-CNN\cite{DBLP:conf/cvpr/GirshickDDM14} started to use the neural network pre-trained on classification dataset for feature extraction. From then on, employing backbone network in detector became a common practice.
Besides CNN, transformer~\cite{DBLP:conf/nips/VaswaniSPUJGKP17} architecture was introduced into vision field. In ViT~\cite{DBLP:conf/iclr/DosovitskiyB0WZ21}, an image is divided into patches and then encoded into a sequence. These attention-based networks~\cite{DBLP:conf/iccv/LiuL00W0LG21,DBLP:conf/iccv/WangX0FSLL0021,DBLP:conf/nips/ZhangY21} also exhibited superior performances in vision tasks. 
Recently, Vim~\cite{DBLP:conf/icml/ZhuL0W0W24} proposes a backbone network based on state space models (SSMs), which serves as a new option for detector design.
To exploit the strength of different architectures, some works~\cite{DBLP:conf/cvpr/Guo0WT00X22,DBLP:conf/cvpr/0003MWFDX22,DBLP:conf/cvpr/XiaWLHS24} devise the hybrid approaches which further boost the dense visual prediction tasks.

No matter whether the detector is CNN-based~\cite{DBLP:conf/iccv/HeGDG17, DBLP:conf/iccv/TianSCH19} or query-based~\cite{DBLP:conf/eccv/CarionMSUKZ20, DBLP:conf/cvpr/SunZJKXZTLYW021}, the detection prediction is derived from the extracted feature. Therefore, in traditional closed-set object detection, backbone networks are supposed to be optimized adequately during training to ensure the detection performance.

Inspired by ViT-CoMer~\cite{DBLP:conf/cvpr/XiaWLHS24}, our method chooses the mixed-architecture which combines ViT and CNN.  
Distinct from these closed-set detection designs, not all modules in our backbone network are trainable. To satisfy the open-vocabulary setting, our backbone network attempts to exploit the information not only from the training data, but also the pre-trained VLM. Thus, a frozen CLIP ViT is employed as a branch of the proposed network, and our feature fusion module performs unidirectional information injection instead of bidirectional interaction.

\textbf{Backbone design for Open-Vocabulary Object Detection.} One of the major issues in OVOD is how to alleviate over-fitting. So far, most OVOD methods still rely on bounding box training to enable the localization ability. However, this attribute inevitably leads to over-fitting on base categories. Another fact is that most OVOD detectors inherit the model structures from the closed-set designs. Backbone network, as the indispensable part of detector, faces the conflict between capturing localization ability and alleviating over-fitting on base-data.

Many works such as~\cite{DBLP:conf/iclr/GuLKC22,DBLP:conf/cvpr/WuZ00L23,DBLP:conf/cvpr/0001LDDLQCL23,DBLP:conf/wacv/0002V024,DBLP:conf/cvpr/ZhaoS0ZKSCM24} choose to optimize the backbone networks with base data for detection, especially these based on the two-stage detectors like~\cite{DBLP:conf/iccv/HeGDG17}. A common strategy is to replace the class-specific module into the class-agnostic and train the entire detector. Usually, the architecture of their backbones is CNN-based. With the additional improvement such as knowledge distillation~\cite{DBLP:conf/iclr/GuLKC22,DBLP:conf/cvpr/0001LDDLQCL23}, pseudo labels~\cite{DBLP:conf/cvpr/ZhaoS0ZKSCM24}, \etc, the detectors could perform well for OVOD task.

There are also some works~\cite{DBLP:conf/iclr/KuoCGPA23,DBLP:conf/cvpr/WuZZL23,DBLP:conf/iclr/WuZX0L0L24} using the frozen backbone networks. Although ViT visual encoder outperforms the ResNet in image-level recognition, the CNN-based network can preserve region attribute better for dense prediction. With modification on the final layer like~\cite{DBLP:conf/eccv/ZhouLD22}, CLIP ResNet could produce the dense feature for detection. Following two-stage detector structure, F-VLM~\cite{DBLP:conf/iclr/KuoCGPA23} freezes its backbone and fine-tunes only the detector head during base training. CORA~\cite{DBLP:conf/cvpr/WuZZL23}, the query-based method, also extracts image features with the frozen CLIP ResNet. To explore the potential of CLIP ViT on dense prediction task, CLIPSelf~\cite{DBLP:conf/iclr/WuZX0L0L24} utilizes a self-distillation strategy, which enhances the recognition ability of CLIP ViT on region-level. Then, F-ViT~\cite{DBLP:conf/iclr/WuZX0L0L24} uses this frozen CLIP ViT as the backbone network to realize the OVOD. 

Unlike these single-branch backbone designs, our method utilizes the two-branch structure which consists of both the trainable CNN and the frozen ViT. Two types of features could be efficiently fused into a stronger representation by the proposed VMC module.
\section{Method}
\label{sec: 2}

\begin{figure*}[htbp]
	\centering{\includegraphics[width=0.95\textwidth]{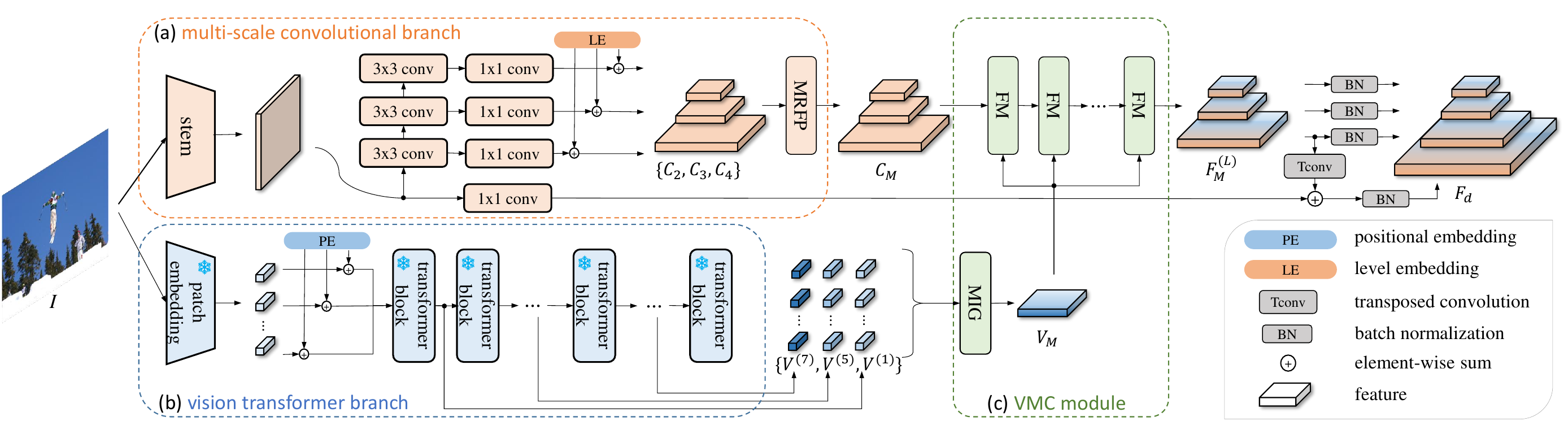}}
	\caption{The overall architecture of proposed VMCNet. Flatten operation is omitted for clarity. Modules marked with snowflake are frozen, the others are optimizable during training. (a) Convolutional branch extracts multi-scale features from the input image. (b) Pre-trained transformer branch provides its intermediate features. (c) VMC module merges the outputs from two branches to generate final multi-scale features.}
	\label{fig:overall}
\end{figure*}

\subsection{Preliminary}
Our detector structure is based on F-ViT~\cite{DBLP:conf/iclr/WuZX0L0L24}. Before detailing our backbone design, we will give a brief description of F-ViT to get a clear view of the entire OVOD method. Unless specified, we choose ViT-B/16 as the default setting for illustration. 

The detection task could be divided into two sub-tasks, \ie, localization and classification. F-ViT follows the two-stage detector design~\cite{DBLP:conf/iccv/HeGDG17}. The RoI features for classification and localization regression are derived from the interpolated intermediate features from the frozen ViT.  We denote the output features of $i$-th transformer block in CLIP ViT as $V^{(i)}$, prediction of the detector could be represented as:
\begin{gather}
F_{RoI} = \phi(V^{(4)}, V^{(6)}, V^{(8)}, V^{(12)}), \\
b_p = {\rm localizer}(F_{RoI}), \\
s_p = {\rm classifier}(F_{RoI}),
\end{gather}
where $b_p$ is the predicted bounding box, $s_p$ is the predicted category score for classification, and $\phi$ denotes the processing includes FPN, RPN, RoIAlign, \etc.

Since the region-level representation of ViT is enhanced with self-distillation, F-ViT also applies the dense feature from last transformer block to improve the classification prediction. Specifically, they use RoI Align to attain the region feature $V_{RoI}$ with RoI bounding box $b_r$. After normalization, the VLM score $s_{VLM}$ is attained via calculating the cosine similarity between $V_{RoI}$ and the encoded text features $T$. The final classification score $s$ is the weighted geometric average of $s_p$ and $s_{VLM}$, where $\beta$ and $\gamma$ are hyper-parameters.
\begin{gather}
V_{RoI} = {\rm RoIAlign}(V^{(12)}, b_r), \\
s_{VLM} = {\rm softmax}(\beta \cdot cos<V_{RoI}, T>), \\
s = s_p ^ \gamma \cdot s_{VLM} ^{1 - \gamma}.
\end{gather}

In our method, we focus on improving the quality of $F_{RoI}$ by using the proposed backbone network. Except the backbone part, we follow the detector design of F-ViT to implement open-vocabulary object detection.

\subsection{Overall Architecture}
Our goal is to design a backbone which could efficiently utilize the information from both the base training data and pre-trained vision transformer. The overall architecture of VMCNet is illustrated in~\cref{fig:overall}, which mainly includes three parts: (a) Multi-Scale Convolutional Branch. (b) Vision Transformer Branch. (c) \textbf{V}iT-Feature-\textbf{M}odulated Multi-Scale \textbf{C}onvolutional (VMC) Module. These three components will be introduced specifically in~\cref{sec: 2_2},~\cref{sec: 2_3} and~\cref{sec: 2_4}, respectively.

The input image $I$ passes through both branches parallelly to obtain the multi-scale convolutional features and vision transformer features.  
Then, two types of features are fed into VMC module, which modulates the multi-scale CNN features with the adapted ViT representations. The output of VMC module is then processed via batch normalization and transposed convolution to attain $F_d$. $F_d$ serves as the backbone features and will be sent to downstream processing. Note that $F_d$ contains the features with resolutions of 1/4, 1/8, 1/16 and 1/32, which is same as the setting in F-ViT.

\subsection{Multi-Scale Convolutional Branch}
\label{sec: 2_2}

Convolutional neural network is well-known for its properties of local continuity and multi-scale capabilities. Besides, it usually have less computation burden than the attention-based network.
Thus, we choose to utilize CNN as the optimizable branch to learn detection representation from base training.
Inspired by ViT-CoMer~\cite{DBLP:conf/cvpr/XiaWLHS24}, we utilize the early stage of its convolutional branch to extract multi-scale convolutional features. As shown in~\cref{fig:overall}, this light-weight version of CNN branch consists of a stem module~\cite{DBLP:conf/cvpr/HeZRS16}, a stack of convolutional layers, level embedding addition and a multi-receptive field feature pyramid (MRFP) module~\cite{DBLP:conf/cvpr/XiaWLHS24}. By using such structure, the additional computation cost is acceptable and we can customize the channel number of output.

Firstly, stem module processes an input image $I$ with the shape of $H \times W \times 3$
into the feature with a resolution reduction of 1/4 of the original image. Then, this intermediate feature goes through three 2-stride $3\times3$ convolutional layers to obtain the features with resolutions of 1/8, 1/16, and 1/32, respectively.

Next, feature at each scale is projected by its corresponding $1\times1$ convolutional layer. Except feature at the largest scale, projected features are added with level embeddings.  The output features are denoted as $C_1 \in \mathbb{R}^{\frac{H}{4} \times \frac{W}{4} \times D} $, $C_2 \in \mathbb{R}^{\frac{H}{8} \times \frac{W}{8} \times D} $, $C_3 \in \mathbb{R}^{\frac{H}{16} \times \frac{W}{16} \times D} $, and $C_4 \in \mathbb{R}^{\frac{H}{32} \times \frac{W}{32} \times D} $, respectively.
For saving the computation cost, outputs excluding $C_1$ are fed into MRFP and VMC module.

MRFP~\cite{DBLP:conf/cvpr/XiaWLHS24} could refine multi-scale features efficiently and expand receptive field. Further details about MRFP may be found in~\cite{DBLP:conf/cvpr/XiaWLHS24}. $C_2$, $C_3$ and $C_4$ are flattened and concatenated into $C \in \mathbb{R}^{(\frac{HW}{8^2} + \frac{HW}{16^2} + \frac{HW}{32^2}) \times D}$. We study the choice of MRFP number in~\cref{sec: num abo}. The output of convolutional branch is attained as follows, where $C_M \in \mathbb{R}^{(\frac{HW}{8^2} + \frac{HW}{16^2} + \frac{HW}{32^2}) \times D}$:
\begin{equation}
C_M = \{C_{M2},C_{M3},C_{M4}\} = {\rm MRFP}(C).
\end{equation}

\subsection{Vision Transformer Branch}
\label{sec: 2_3}
Pre-trained on an enormous number of image-text pairs, visual encoder in CLIP~\cite{DBLP:conf/icml/RadfordKHRGASAM21} could process an image into the encoded feature aligned with text feature space. CLIPSelf~\cite{DBLP:conf/iclr/WuZX0L0L24} further improves the visual transformer encoder with self-distillation for dense prediction task.
Thus, we use the frozen visual encoder from~\cite{DBLP:conf/iclr/WuZX0L0L24} as ViT branch to attain the modulating feature.

In ViT branch, input image $I$ firstly passes through a large kernel $16 \times 16$ convolutional layer (patch embedding). After added with the positional embedding, we attain the input feature $V^{(0)} \in \mathbb{R}^{\frac{H}{16} \times \frac{W}{16} \times D}$. $V^{(0)}$ is fed into a series of vision transformer blocks to attain the intermediate features.

To make full use of the representation ability of the frozen ViT, we collect multiple intermediate features from this branch. Under ViT-B/16 setting, we choose $V^{(1)}$, $V^{(5)}$ and $V^{(7)}$ for feature modulation. Study about the choice of ViT layers is presented in~\cref{sec: vit layer}. 
Since not all transformer blocks are used to attain the intermediate features, we could skip the computation of remaining blocks during training. In inference phase, we apply the whole vision transformer to generate feature map from the last block for dense prediction.

\subsection{VMC Module Design}
\label{sec: 2_4}

Inspired by ViT-CoMer~\cite{DBLP:conf/cvpr/XiaWLHS24}, a strong backbone network utilizing the pretrained ViT via feature interaction, we design the VMC module based on its CNN-Transformer bidirectional fusion interaction (CTI) block. As shown in~\cref{fig:overall}, VMC module mainly consists of a modulating information generation (MIG) block and $L$ feature modulation (FM) blocks.
Notice that there are several crucial differences between ViT-CoMer and our method. 

First, VMC module focuses on injecting the information from a sophisticated visual encoder into CNN features, which is unidirectional instead of bidirectional. To keep the rich knowledge in visual encoder, the vision transformer in VMCNet is frozen while ViT-CoMer is fine-tunable.
Moreover, VMC module is applied merely at the end of two branches, which owns different structure from theirs. Our method is designed for OVOD task, while ViT-CoMer aims to enhance the closed-set object detector.

\begin{figure}[t]
	\centering{\includegraphics[width=0.45\textwidth]{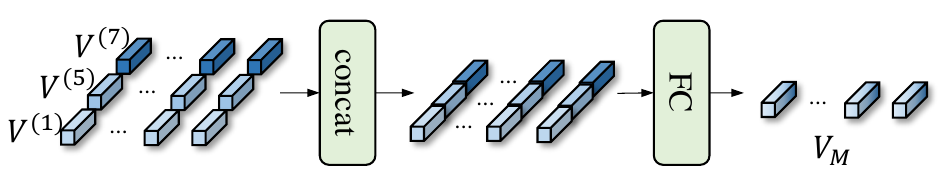}}
	\caption{Structure of modulating information generation module. FC denotes the fully connected layer.}
	\label{fig:mig}
\end{figure}

\textbf{Modulating Information Generation.} We attempt to utilize the ViT features collected from different blocks to generate a unified representation. Thus, we propose a novel module named as modulating information generation. Its structure is illustrated in~\cref{fig:mig}.
Firstly, the ViT features $V^{(1)}$, $V^{(5)}$ and $V^{(7)}$ are concatenated along the channel dimension. The concatenated high-dimensional features contains representations from different depths in ViT. Then, a linear projection layer is applied, which not only reduces the channel number from $3*D$ to $D$, but also simply adapts the ViT features to the downstream processing. We implement this linear projection with a fully connected layer. Via MIG, we could attain the adaptive modulating feature. The processing of MIG could be denoted as:
\begin{equation}
V_M={\rm FC}({\rm concat}(V^{(1)}, V^{(5)}, V^{(7)})).
\end{equation}

\textbf{Feature Modulation.} With the adaptive modulating features from MIG and the multi-scale features from CNN branch, we achieve the feature modulation by stacking a serial of feature modulation (FM) blocks in a cascaded manner as shown in~\cref{fig:overall}. Output of the previous FM block is sent to the next one to achieve progressive feature refinement. All blocks use the same modulating feature $V_M$. 

\begin{figure}[t]
	\centering{\includegraphics[width=0.45\textwidth]{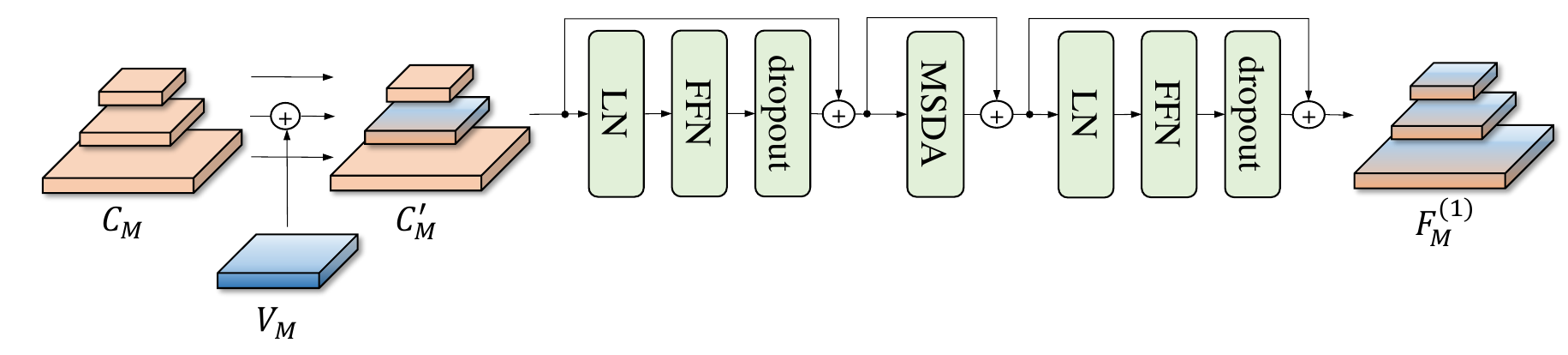}}
	\caption{Structure of feature modulation block. We illustrate processing in the first block as example.}
	\label{fig:fm}
\end{figure}

FM block shares the same structure as ViT-to-CNN (VtoC) block in ViT-CoMer, its structure is illustrated in~\cref{fig:fm}. This block consists of a multi-scale deformable attention (MSDA)~\cite{DBLP:conf/iclr/ZhuSLLWD21}, two layer normalization (LN) operations, two feed-forward networks (FFN), two dropout layers and three skip connection sums.

In first FM block, $V_M$ is added into the CNN feature at 1/16 scale. We denote the new multi-scale features as $C_M^{\prime}$. Then $C_M^{\prime}$ goes through a layer normalization layer, a feed-forward network and a dropout layer, the processing is represented as below:
\begin{gather}
C_M^{\prime} = \{C_{M2},C_{M3}+V_M,C_{M4}\}, \\
C_F = {\rm dropout}({\rm FFN}({\rm LN}(C_M^{\prime}) )) + C_M^{\prime}.
\end{gather}

Although the element-wise sum operation simply fuses the features, the representation from ViT is merely injected into the features at second scale.  
As we know, deformable attention (DA) utilizes the reference points to sample the features to attain the values. In MSDA, each scale will provide a fixed number of reference points, that is, ViT representation could be scattered into the features at all scales via MSDA. The output of MSDA is then processed similarly as the above:
\begin{gather}
C_F^{\prime} = {\rm MSDA}(C_F) + C_F, \\
F_M^{(1)} = {\rm dropout}({\rm FFN}({\rm LN}(C_F^{\prime}) )) + C_F^{\prime}.
\end{gather}

Through the operations in first FM block, features at each scale of $F_M^{(1)}$ has carried the representations from frozen ViT. In the next block, we repeat the processing as illustrated above to further inject the ViT representations into the multi-scale features. For $i$-th ($i\geq2$) block, feature modulation is denoted as:
\begin{equation}
F_M^{(i)} = \{F_{M2}^{(i)}, F_{M3}^{(i)}, F_{M4}^{(i)}\} = {\rm FM}(F_M^{(i-1)}, V_M).
\end{equation}

We also study the choice of $L$ in~\cref{sec: num abo}. The output of final block is denoted as $F_M^{(L)}$, we interpolate the feature with largest scale in $F_M^{(L)}$ and add it with $C_1$. After batch normalization, the four-scale features $F_d$ could be used as the backbone features for downstream steps. The processing for $F_M^{(L)}$ is represented as below:
\begin{align}
F_{d1}&={\rm BN}({\rm Tconv}(F_{M2}^{(L)}) + C_1), \\
F_{di}&={\rm BN}(F_{Mi}^{(L)}), i=2,3,4, \\
F_d &= \{F_{d1}, F_{d2}, F_{d3}, F_{d4}\}.
\end{align}
\section{Experiments}

\subsection{Experiment Setup}
\subsubsection{Datasets and evaluation metrics}
Our method is evaluated on two popular open-vocabulary  object detection datasets, OV-COCO~\cite{DBLP:conf/eccv/LinMBHPRDZ14} and OV-LVIS~\cite{DBLP:conf/cvpr/GuptaDG19}. 

\textbf{OV-COCO.}
MSCOCO is one of the most commonly used datasets in object detection. To evaluate a detector under open-vocabulary setting, OVR-CNN~\cite{DBLP:conf/cvpr/ZareianRHC21} proposes to build the base set with 48 categories of its data, and the other 17 categories of data are used as the novel. We follow this rule to train our method with the base set data and evaluate it with the set including both the base and the novel set. As the behaviors of existing works, bounding box AP at IoU threshold 0.5 of novel categories is employed as the main metric, which is denoted as AP$_{50}^{\mathrm{novel}}$.

\textbf{OV-LVIS.}
LVIS is a large-scale dataset for instance segmentation, which contains 1203 categories of objects. The category set is split into rare (337 categories), common (461) and frequent (405) based on the number of images the objects appear. We follow ViLD~\cite{DBLP:conf/iclr/GuLKC22} to split the dataset. Rare categories are split into the novel while common and frequent categories are split into the base. The mask AP averaged on IoUs from 0.5 to 0.95 of rare categories, denoted as mAP$_{r}$, is chosen as the main metric for evaluation.

\subsubsection{Implementation Details}
Our detection framework is based on F-ViT~\cite{DBLP:conf/iclr/WuZX0L0L24}. We use the self-distilled visual encoder from CLIPSelf to initialize the weights of vision transformer branch. For OV-COCO, we use the CLIPSelf with proposal distillation setting and leverage three ViT layers. For OV-LVIS, CLIPSelf with the patch distillation setting is used and five ViT layers are employed. 
We conduct the experiments with 8 RTX 3090 GPUs and set the batch size as 1 on each GPU. For fair comparisons, our method inherits the configurations of baseline except the proposed part. Following the implementation of F-ViT, we use the same training scheme.  AdamW optimizer with a learning rate of $1.25\times10^{-5}$ is utilized, detector for OV-COCO is trained for 3 epochs while 48-epoch training is for OV-LVIS.

\subsection{Quantitative Comparisons}
\begin{table}
	\centering
		\scalebox{.9}{
			\begin{tabular}{l|c|c}
				\toprule
				Method& Backbone&AP$_{50}^{\mathrm{novel}}$(\%)\\ \hline
				ViLD~\cite{DBLP:conf/iclr/GuLKC22}  & RN50& 27.6\\
				Detic~\cite{DBLP:conf/eccv/ZhouGJKM22}&RN50 & 27.8 \\
				F-VLM~\cite{DBLP:conf/iclr/KuoCGPA23}& RN50 &28.0\\
				OV-DETR~\cite{DBLP:conf/eccv/ZangLZHL22} & RN50 & 29.4\\
				BARON-KD~\cite{DBLP:conf/cvpr/WuZ00L23}& RN50  & 34.0\\
				CLIM BARON~\cite{DBLP:conf/aaai/WuZX00L24}& RN50  & 36.9\\
				SAS-Det~\cite{DBLP:conf/cvpr/ZhaoS0ZKSCM24}& RN50-C4  & 37.4\\
				EdaDet~\cite{DBLP:conf/iccv/ShiY23b}& RN50  & 37.8\\
				SIA~\cite{DBLP:conf/mm/WangZXP24}&RN50x4& 41.9\\
				CORA+~\cite{DBLP:conf/cvpr/WuZZL23}&RN50x4& 43.1\\
				\hline
				RO-ViT~\cite{DBLP:conf/cvpr/KimAK23} & ViT-L/16 &  33.0 \\
				CFM-ViT~\cite{DBLP:conf/iccv/KimAK23a} & ViT-L/16 & 34.1\\
				
				BIND~\cite{DBLP:conf/cvpr/ZhangZZZGLX24} & ViT-L/16 & 41.5\\ 
				
				\hline
				F-ViT+CLIPSelf~\cite{DBLP:conf/iclr/WuZX0L0L24} & ViT-B/16& 37.6\\
				F-ViT+CLIPSelf \dag ~\cite{DBLP:conf/iclr/WuZX0L0L24} & ViT-B/16& 38.7\\
				F-ViT+CLIPSelf~\cite{DBLP:conf/iclr/WuZX0L0L24} & ViT-L/14 &44.5\\
				F-ViT+CLIPSelf \dag ~\cite{DBLP:conf/iclr/WuZX0L0L24} & ViT-L/14 &44.5\\
				\hline
				F-ViT+VMCNet (Ours) & VMCNet-B & 44.3 (+5.6)\\ 
				F-ViT+VMCNet (Ours) & VMCNet-L & \textbf{48.5} (+4.0)\\
				\bottomrule
			\end{tabular}
		}
		\caption{Comparison with state-of-the-art methods on OV-COCO benchmark. `\dag' ~denotes that this result is obtained from the reimplemented experiments under our local environment.}
		\label{tab: 1}
\end{table}

\textbf{OV-COCO.}
We report the comparison with previous methods in~\cref{tab: 1}. VMCNet based on ViT-B/16 is denoted as VMCNet-B while VMCNet-L refers to the one based on ViT-L/14. To get rid of the influence from experiment environment, we train the baseline method under our local environment and display the results. We choose the highest performance of baseline method for comparison.  Under ViT-B/16 setting, our method achieves 44.3 AP$_{50}^{\mathrm{novel}}$, which is even comparable to the baseline with ViT-L/14.
The performance can be further improved when increasing the scale of network. F-ViT equipped with VMCNet-L achieves 48.5 AP$_{50}^{\mathrm{novel}}$ and also outperforms the baseline method by an obvious margin.
Through the evaluation on OV-COCO, we found VMCNet could effectively boost the detection performances on novel categories.

\textbf{OV-LVIS.}
\Cref{tab: 2} lists the performances of state-of-the-art methods on OV-LVIS. VMCNet also surpasses the previous methods. Compared to the baseline with ViT-B/16, our method can bring +2.3 mAP$_{r}$ gain. Under ViT-L/14 setting, VMCNet can outperform the baseline approach by 3.2 mAP$_{r}$. 
Therefore, VMCNet is proved to be effective on detecting objects of novel categories.

\begin{table}
	\centering
		\scalebox{.9}{
			\begin{tabular}{l|c|c}
			\toprule
			Method& Backbone & mAP$_{r}$(\%) \\ \hline
			ViLD~\cite{DBLP:conf/iclr/GuLKC22} & RN50 & 16.6 \\
			OV-DETR~\cite{DBLP:conf/eccv/ZangLZHL22} & RN50 & 17.4 \\
			BARON-KD~\cite{DBLP:conf/cvpr/WuZ00L23} & RN50 & 22.6\\
			EdaDet~\cite{DBLP:conf/iccv/ShiY23b}& RN50  & 23.7\\
			CORA+~\cite{DBLP:conf/cvpr/WuZZL23}& RN50x4 & 28.1 \\
			SAS-Det~\cite{DBLP:conf/cvpr/ZhaoS0ZKSCM24}& RN50x4-C4 & 29.1 \\
			F-VLM~\cite{DBLP:conf/iclr/KuoCGPA23}& RN50x64& 32.8\\ 
			\hline
			OWL-ViT~\cite{DBLP:journals/corr/abs-2205-06230} & ViT-L/14 & 25.6 \\ 
			RO-ViT~\cite{DBLP:conf/cvpr/KimAK23} & ViT-L/16 & 32.4 \\ 
			BIND~\cite{DBLP:conf/cvpr/ZhangZZZGLX24} & ViT-L/16 & 32.5 \\ 
			CFM-ViT~\cite{DBLP:conf/iccv/KimAK23a} & ViT-L/16 & 33.9 \\ 
			RO-ViT~\cite{DBLP:conf/cvpr/KimAK23} & ViT-H/16 & 34.1 \\ 
			\hline
			F-ViT+CLIPSelf~\cite{DBLP:conf/iclr/WuZX0L0L24} & ViT-B/16 & 25.5 \\
			F-ViT+CLIPSelf~\cite{DBLP:conf/iclr/WuZX0L0L24} & ViT-L/14 & 35.2 \\ 
			\hline
			F-ViT+VMCNet (Ours) & VMCNet-B  & 27.8(+2.3)\\
			F-ViT+VMCNet (Ours) & VMCNet-L & \textbf{38.4}(+3.2)\\
			\bottomrule
		\end{tabular}
		}
		\caption{Comparison with state-of-the-art methods on OV-LVIS.}
		\label{tab: 2}
\end{table}

\subsection{Ablation Experiments}

We conduct ablation studies on OV-COCO benchmark under ViT-B/16 setting. Unless specified, we report performances with the highest AP$_{50}^{\mathrm{novel}}$, the default setting is the same as the statements in~\cref{sec: 2}.

\begin{table}
	\begin{minipage}{0.5\textwidth}
	\centering
		\scalebox{0.7}{
			\begin{tabular}{ccccc|ccc}
				\hline
				ViT & CNN & MIG & FM*  & FM  & AP$_{50}^{\mathrm{novel}}$(\%) &\textcolor{gray}{AP$_{50}^{\mathrm{base}}$(\%) }& TParams\\
				\hline
				\ding{51} & \ding{53} & \ding{53} & \ding{53} & \ding{53} & 38.70  & \textcolor{gray}{53.94} & 21.02M\\
				\ding{51} & \ding{51} & \ding{53} & \ding{53} & \ding{53} & 18.40  & \textcolor{gray}{13.89} &  18.50M \\
				\ding{51} & \ding{51}  & \ding{51} & \ding{51} & \ding{53} & 23.06  & \textcolor{gray}{15.01} & 24.86M \\
				\ding{51} & \ding{51}  & \ding{51} & \ding{53} & \ding{51} &  \textbf{44.33}   & \textcolor{gray}{49.93}& 24.86M \\
				\hline
			\end{tabular}
		}	
		\caption{Ablation study on main components of VMCNet. `FM*' refers to the FM block \textbf{without} $V_M$ feature addition. `TParams' means the number of trainable parameters in the whole detector.}
		\label{tab: 3}
	\end{minipage}
\end{table}
\subsubsection{Effectiveness of model structure}
\label{sec: num abo}
In this part, we attempt to analyze the effectiveness of components in VMCNet and evaluate the feasibility of these structures.  

The ablation study on the main components is shown in~\cref{tab: 3}. We start from the baseline method which only applies the CLIP ViT, its performance is reported in first row. In second row, backbone is replaced by the trainable CNN, ViT only serves as the RoI classifier and the last few transposed convolutional layers to interpolate its features are removed. This design corresponds to the illustration in~\cref{fig:intro_b}, where detector could only rely on the features extracted by the simple CNN and thus have poor performance. In third row, we further add the abridged FM blocks which remove the $V_M$ addition, though the performance is improved by a little, detector still suffers from lack of ViT representation. In the bottom line, the entire VMCNet is implemented. We can see that the increase on parameter cost is lower than 4M. Therefore, these main components can effectively utilize the information from both the frozen ViT and base training data. 

\begin{table}
	\begin{minipage}{0.5\textwidth}
		\centering
		\scalebox{1.2}{
		\begin{tabular}{c|cc}
			\hline
			$L$ & AP$_{50}^{\mathrm{novel}}$(\%) &\textcolor{gray}{AP$_{50}^{\mathrm{base}}$(\%) }\\
			\hline
			$1\times2$  & 43.11 &  \textcolor{gray}{49.42}\\
			$1\times3$  & \textbf{44.33} &  \textcolor{gray}{49.93}\\
			$1\times4$  & 43.95 &  \textcolor{gray}{50.12}\\
			$2\times3$  & 43.65 &  \textcolor{gray}{50.47}\\
			$3\times1$  & 43.85 &  \textcolor{gray}{50.30}\\
			
			\hline
		\end{tabular}
		}
		\caption{Ablation study on the strategy of placing FM blocks.`$s\times t$' represents that there are $s$ groups of FMs, each group is supported by a MIG module and contains $t$ FM blocks. }
		\label{tab: 4}
	\end{minipage}
\end{table}

\begin{figure}
	\centering
	\subfloat[$1 \times 3$]{\includegraphics[width=0.1\textwidth,height=0.165\linewidth]{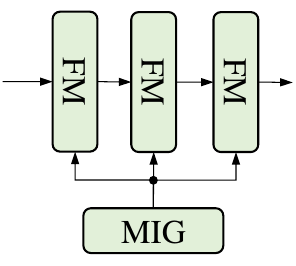}\label{fig: abo_fm_stru_a}}\hfill
	\subfloat[$2 \times 3$]{\includegraphics[width=0.2\textwidth,height=0.165\linewidth]{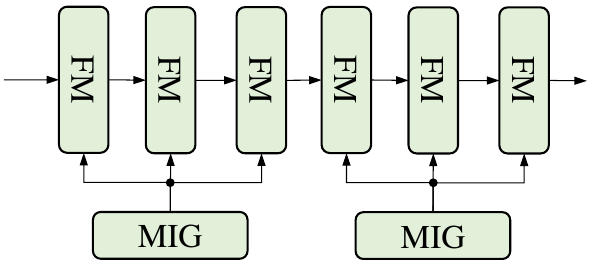}\label{fig: abo_fm_stru_b}}\hfill
	\subfloat[$3 \times 1$]{\includegraphics[width=0.1\textwidth,height=0.20\linewidth]{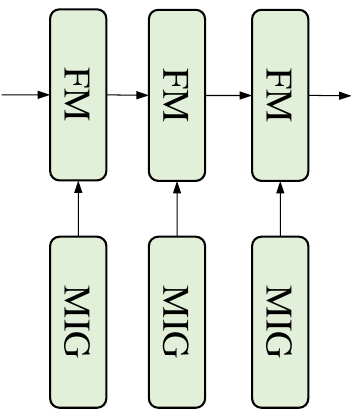}\label{fig: abo_fm_stru_c}}
	\caption{Optional strategies of placing FM blocks in VMC module.}
	\label{fig: abo_fm_stru}
\end{figure}

Besides, we explore the strategy of placing FM blocks in VMC modules. As shown in~\cref{tab: 4}, we present 5 strategies. We can see that the highest AP$_{50}^{\mathrm{novel}}$ is achieved in the second row, that is, the optimal value of $L$ is 3. To better understand these strategies, the local designs in 2-nd, 4-th and 5-th row are illustrated in~\cref{fig: abo_fm_stru_a},~\cref{fig: abo_fm_stru_b} and~\cref{fig: abo_fm_stru_c}, respectively. Compared to the 2-nd row, the 4-th adds a repeated structure behind, which does not improve performance. In 5-th line, each FM block is supported by a different MIG module. This result shows using more MIG modules can not bring gain, either.
Therefore, we choose the strategy in second row for OV-COCO setting.

We also analyze the impact of the number of MRFP modules in CNN branch.  The first row in~\cref{tab: 5} corresponds to the structure removing MRFP from VMCNet, which leads to a drop of 1.4 AP$_{50}^{\mathrm{novel}}$. 
The bottom line is the result of cascading two MRFP modules, which shows that increasing $N_{MRFP}$ does not improve the performance. Thus, only one MRFP module is used in our CNN branch design.

\begin{table}
	\centering
		\scalebox{1.}{
		\begin{tabular}{c|cc}
			\hline
			$N_{MRFP}$ & AP$_{50}^{\mathrm{novel}}$(\%) & \textcolor{gray}{AP$_{50}^{\mathrm{base}}$(\%) }\\
			\hline
			0  & 42.93  & \textcolor{gray}{49.71}\\ 
			1  & \textbf{44.33}  & \textcolor{gray}{49.93}\\ 
			2  & 42.99  & \textcolor{gray}{49.55}\\ 
			\hline
		\end{tabular}
		}
		\caption{Effectiveness of $N_{MRFP}$. `$N_{MRFP}$' denotes the number of MRFP modules.}
		\label{tab: 5}
\end{table}

\begin{figure*}[t]
	\centering
	\subfloat[Visual results from baseline method.]{
	\includegraphics[width=0.2\textwidth,height=0.135\linewidth]{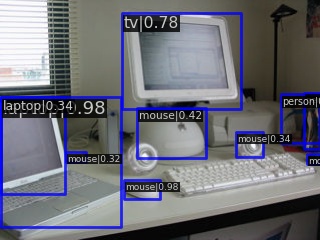}
	\hspace{0.02\textwidth}
	\includegraphics[width=0.2\textwidth,height=0.135\linewidth]{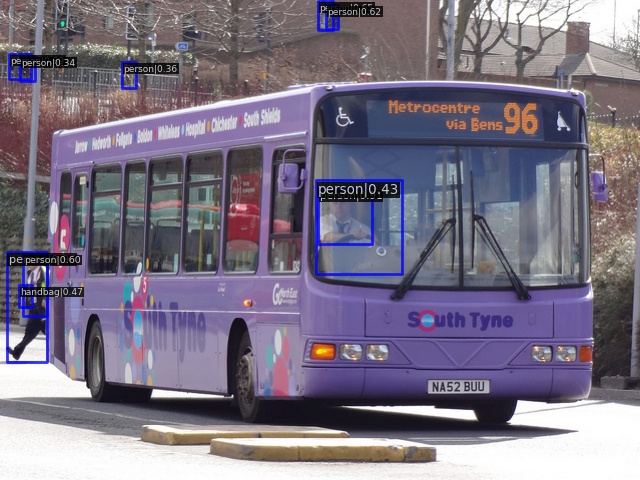}
	\hspace{0.02\textwidth}
	\includegraphics[width=0.2\textwidth,height=0.135\linewidth]{pic_material/visual_resuts/collected_vrs/base_vrs/000000005477.jpg}
	\hspace{0.02\textwidth}
	\includegraphics[width=0.2\textwidth,height=0.135\linewidth]{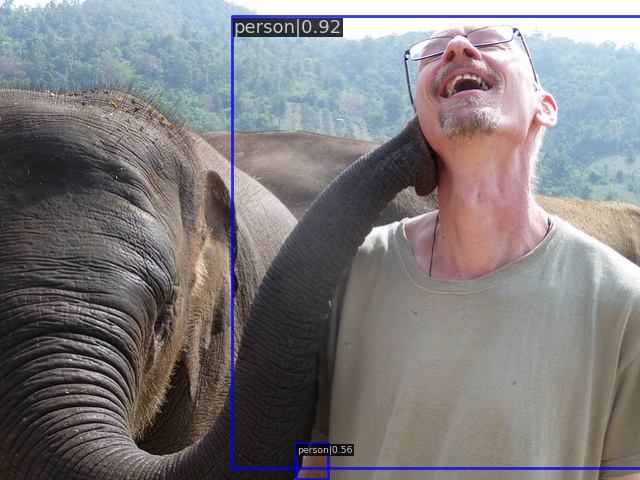}
	}
	\\
	\subfloat[Visual results from our method.]{
	\includegraphics[width=0.2\textwidth,height=0.135\linewidth]{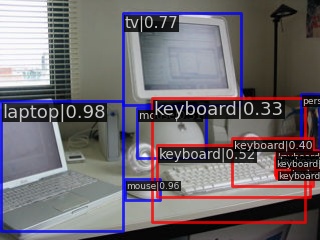}
	\hspace{0.02\textwidth}
	\includegraphics[width=0.2\textwidth,height=0.135\linewidth]{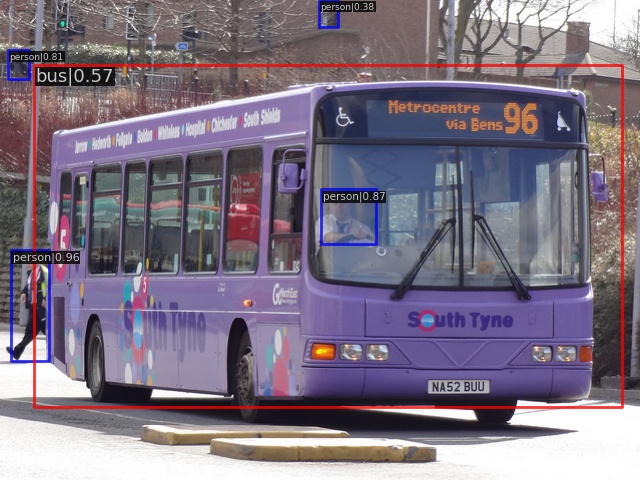}
	\hspace{0.02\textwidth}
	\includegraphics[width=0.2\textwidth,height=0.135\linewidth]{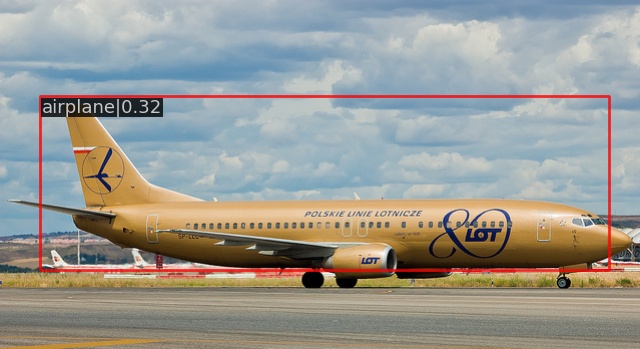}
	\hspace{0.02\textwidth}
	\includegraphics[width=0.2\textwidth,height=0.135\linewidth]{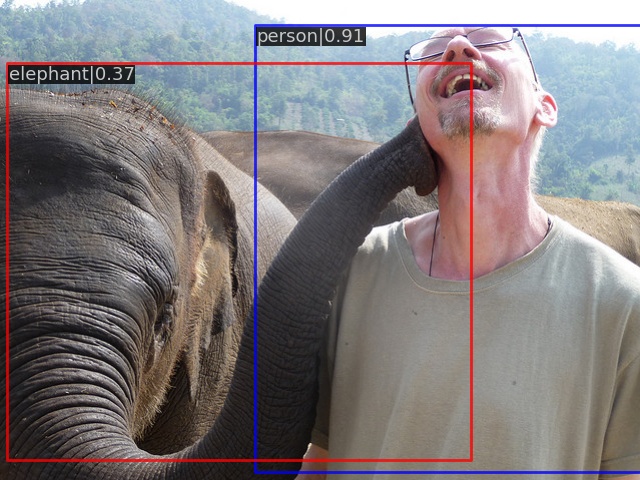}
	}
	
	\caption{Comparison of visual results. The detection results from baseline are displayed in the upper line while ours are in the bottom line. Both the detector architectures are based on ViT-B/16. The image samples are collected from COCO validation set. Blue bounding box represents that the predicted category belongs to the base while red box belongs to the novel.}
	\label{fig: vr}
\end{figure*}

\subsubsection{Effectiveness of ViT branch}
\label{sec: vit layer}

In this part, we conduct ablation experiments on the used ViT feature layers as shown in~\cref{tab: 6}. 
The default setting we use is the 4-th row. 
In first three rows, one of the default feature layers is removed. As a result, these results are inferior to the setting in 4-th line. However, in 5-th row, the detection performance on novel categories degenerates obviously when the extra ViT feature layer $V^{(9)}$ is added. In the last row, features from the deep blocks of ViT are applied, which leads to the poorer performance. Thus, we find that the ability to detect novel categories may benefit from features of some certain layers of frozen ViT. In OV-COCO setting, features from the high level of ViT may even hurt the performances.

At last, there still remains a doubt whether the gain on performance is due to the transformer model structure or the knowledge of pre-training. To find out the answer, we unfreeze the ViT branch and use an extra frozen ViT for RoI classification. The results are presented in~\cref{tab: 7}. The second row shows that using the trainable ViT branch leads to more parameter cost and does not enhance the performance. Therefore, the proposed method can utilize the knowledge of pre-training to improve the detection performances.

\begin{table}
	\centering
		\scalebox{.9}{
			\begin{tabular}{c|cc}
				\hline
				ViT layers & AP$_{50}^{\mathrm{novel}}$(\%) & \textcolor{gray}{AP$_{50}^{\mathrm{base}}$(\%) } \\
				\hline
				$V^{(1)}$,$V^{(5)}$ & 39.73 & \textcolor{gray}{42.29} \\
				$V^{(1)}$,$V^{(7)}$ & 43.25 & \textcolor{gray}{49.16} \\
				$V^{(5)}$,$V^{(7)}$ & 43.47 & \textcolor{gray}{49.93} \\
				\hline
				$V^{(1)}$,$V^{(5)}$,$V^{(7)}$ & \textbf{44.33} & \textcolor{gray}{49.93} \\
				\hline
				$V^{(1)}$,$V^{(5)}$,$V^{(7)}$,$V^{(9)}$ & 36.02 & \textcolor{gray}{48.61} \\
				$V^{(6)}$,$V^{(10)}$,$V^{(12)}$ & 33.59 & \textcolor{gray}{49.47} \\
				\hline
			\end{tabular}
		}
		\caption{Effectiveness of the used ViT feature layers.}
		\label{tab: 6}
\end{table}

\begin{table}
	\centering
		\begin{tabular}{c|ccc}
			\hline
			ViT state   & AP$_{50}^{\mathrm{novel}}$(\%) & \textcolor{gray}{AP$_{50}^{\mathrm{base}}$}(\%) & TParams\\
			\hline
			frozen   & \textbf{44.33} & \textcolor{gray}{49.93} & 24.86M \\
			trainable & 38.75 & \textcolor{gray}{47.59} & 75.65M \\ 
			
			\hline
		\end{tabular}
		\caption{Frozen ViT branch $vs$ trainable ViT branch.}
		\label{tab: 7}
\end{table}

\subsection{Qualitative Visualization}
Detection results of the baseline method and ours are visualized in~\cref{fig: vr}. We set the score threshold as 0.3 to filter the predicted bounding box whose classification score is lower than this value. From the visual comparisons, we observe that our method is more confident to  recognize the novel objects. For example, in second column, the baseline fails to detect the bus in image, while ours could predict it with a relatively high confidence score.

\section{Conclusion}

In this paper, we propose a two-branch backbone network for OVOD task, named as VMCNet. Our method combines the advantages of CNN and ViT. The frozen ViT branch saves the generalization ability while the trainable CNN learns information of the base training data. At last, the proposed VMC module could modulate multi-scale convolutional features with the representations from ViT branch.
Our design provides an effective scheme to utilize the knowledge from both the pre-training and base training.
As a result, VMCNet improves the detection performances on novel categories effectively and efficiently.  
Experimental results demonstrate that our method improves the baseline significantly and outperforms the state-of-the-art methods.

{\small
\bibliographystyle{ieee_fullname}
\bibliography{egbib}
}

\clearpage

\end{document}